# Few-shot Object Detection with Feature Attention Highlight Module in Remote Sensing Images

Zixuan Xiao[a], Ping Zhong*[a], Yuan Quan[a], Xuping Yin[a], Wei Xue[ab]

[a]National Key Laboratory of Science and Technology on Automatic Target Recognition, College of Electrical Science, National University of Defense Technology, Changsha, China;
[b]School of Computer Science and Technology, Anhui University of Technology, Maanshan, China

**ABSTRACT**

In recent years, there are many applications of object detection in remote sensing field, which demands a great number of labeled data. However, in many cases, data is extremely rare. In this paper, we proposed a few-shot object detector which is designed for detecting novel objects based on only a few examples. Through fully leveraging labeled base classes, our model that is composed of a feature-extractor, a feature attention highlight module as well as a two-stage detection backend can quickly adapt to novel classes. The pre-trained feature extractor whose parameters are shared produces general features. While the feature attention highlight module is designed to be light-weighted and simple in order to fit the few-shot cases. Although it is simple, the information provided by it in a serial way is helpful to make the general features to be specific for few-shot objects. Then the object-specific features are delivered to the two-stage detection backend for the detection results. The experiments demonstrate the effectiveness of the proposed method for few-shot cases.

## 1. INTRODUCTION

Thanks to the development recently in computer vision, the past few decades have seen the rapid progress in remote sensing technology which has brought quantities of applications[1], such as forecasts of disasters and assistance in rescue operations. Among these applications, object detection has played an important part thanks to the deep convolutional neural networks (CNNs). While as we all know, CNNs based methods rely heavily on huge amounts of training data. However, in some cases, the training data is not easy to get. Therefore, the number of training data is rare in these cases, which may lead to bad performance as CNNs severely overfit and fail to generalize. This may be even worse for the complex object detection.

The research which targets at the case that the number of training data is extremely rare is called few-shot learning[2] (FSL). In this paper, we focus on the challenging task, i.e. few-shot object detection. In detail, we aim to make the model able to detect the novel objects when there are some base classes with sufficient number of samples and novel classes with only a few samples. In FSL field, the research[3] mainly focuses on the classification task that is much easier compared with object detection task. Essentially, this is because the object detection task should not only tell us which class the object belongs to as what the classification task does, but also tell us where it is in the image. On the other hand, when meeting few-shot cases, the data available in classification is a few images of the class. While in the detection task, only a few annotations of the class are available, which is quite different from the former. As a consequence, it is not feasible to apply the methods that aim to solve few-shot classification task to the few-shot object detection task.

Our proposed model consists of a feature extractor, a feature attention highlight module and a two-stage detection backend. In our model, the parameters of the first part are pre-trained and the networks used in the feature attention highlight module are designed to be light-weighted and simple. All of these are designed to fit the few-shot situation. As the way some meta-learning problems[4] and some few-shot detection works[5] did to split training data, we split them into support set as well as query set. In query set, the images are from the detection dataset, while in support set, the images are parts of objects cropped from the same detection dataset. In a single training step, the inputs are a combination of a query image and some support images. In detail, a query image is taken by the feature extractor as input to obtain general features. Simultaneously, the support images are fed into the feature attention highlight module. After that we can get so-called feature-exciting factors which contain the information from the few-shot objects in a serial form. These factors are implied to highlight the few-shot object features from coarse to fine. Eventually, the object-specific features are fed into the detector for results. To make sure that our model can quickly adapt to few-shot cases, we train it according to a two-phase training scheme like what transfer learning do.

*corresponding author: zhongping@nudt.edu.cn

The main contributions of this paper are as follows. We concentrate on the challenging few-shot object detection that is of great practical value, while is less explored than classification in few-shot learning field. We design a few-shot object detector which is made up of two parts and is able to detect novel classes based on only a few data. We study the problem of few-shot object detection in remote sensing field which can be utilized for future applications.

## 2. RELATED WORKS

Object detection is a rather challenging task compared with classification task in computer vision. The methods of it can be divided into two categories with the help of CNN: proposal-based and proposal-free. These two categories are also commonly defined as two-stage and one-stage. RCNN series detectors[6-7] fall into the first category, all of which extract proposals at the beginning through CNNs such as region-proposal-network (RPN) and then aim to detect those proposals. In contrast, YOLO series detectors[8-9] or SSD series[10] detectors can detect objects through a single convolutional network. They are fast but not as accurate as the first category.

Most works in object detection field focus on the cases where training data is sufficient. However, there exist situations where training data is extremely rare. The research that focuses on learning from only a few training examples is called few-shot learning. Recently, great progress about classification tasks has been made in few-shot cases. The popular solutions are mainly based on transfer learning or meta-learning[11-12]. However, the challenging few-shot object detection[13] task is far from fully researched.

In remote sensing field, there are many works[14-15] about object detection. Almost all of them are based on the situation where there is sufficient training data. The attention of those works is on how to make the detection networks adapt to remote sensing input. There also exist some works[16] on few-shot classification in remote sensing field. However, as for the few-shot object detection, the study[17] is rather fewer. In our work, with the assistance of ideas of transfer learning, we aim at the challenging task in remote sensing field.

## 3. PROPOSED METHOD

In many application scenarios, there exists data from some classes which is easy to get and thus in a large number. We call these classes base classes. On the contrary, the new classes of which the data is quite rare or cannot easily get are novel classes. In these scenarios, we aim to obtain a few-shot detector which can be trained to detect novel object through fully leveraging knowledge from base classes.

As shown in Figure 1, our few-shot detector consists of a feature extractor, an attention highlight module and a detection backend. The feature extractor takes a query image as input for the general feature maps. Simultaneously, a group of support images from every class (only one from each class) are processed by the weight-shared feature extractor to get embeddings of these images. Then they are fed into two highlighters which are in a serial way. After that, two groups of feature-exciting factors each of which contains the information of the object from every class are produced to do deep-wise cross correlation with the general feature maps to get object-specific features. In a serial way, the factors are from coarse to fine and the general features are processed step by step to be specific. Eventually, the features are fed into the detection backend to obtain the final results and the results are fused ultimately.

### 3.1 Feature Attention Highlight Module

There are two parts in our feature attention highlight module. The first part is the feature extractor and the other part is composed of two feature highlighters. The feature extractor shares its weights with the backbone of the detection model. In every iteration, the inputs of it are a group of support images. The classes of them are consistent with the classes of the objects that are to be detected. Particularly, there is only one support image in a group for a single class. The support images are all resized to be 224 × 224 for simplicity. Because the weights are shared, there is no additional parameter that is to be optimized.

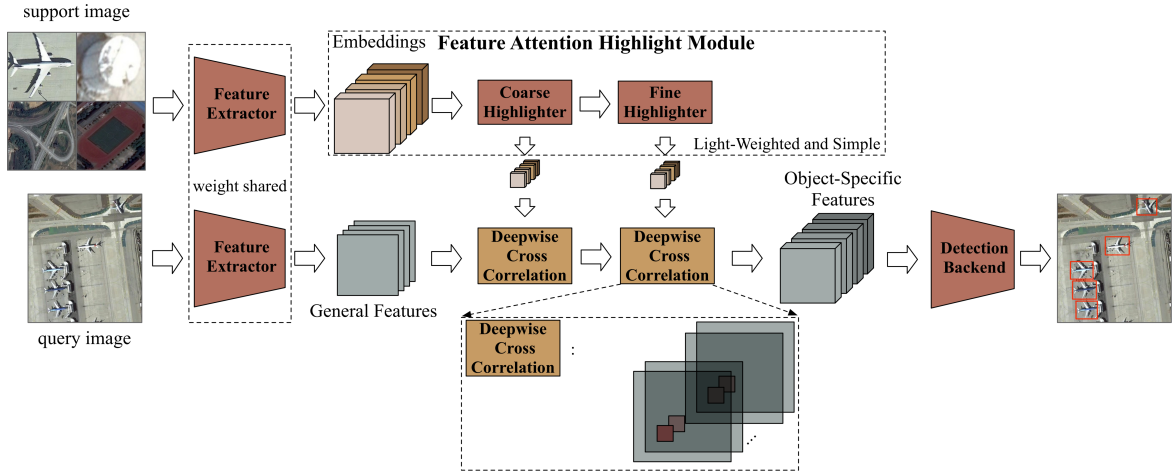

Figure 1. The architecture of our proposed few-shot detector.

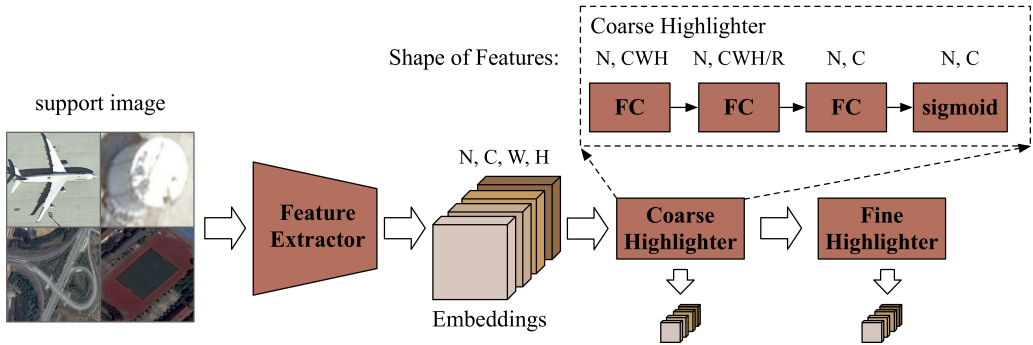

Figure 2. The architecture of feature attention highlight module.

The architecture of feature highlighters is in Figure 2. The first highlighter, as we call it coarse highlighter, directly take embeddings as input. Specifically, the embeddings are fed into three fully-connected layers to decrease the number of channels in order to reduce parameters and to match the channels of the feature maps of the query image. Eventually, through a sigmoid layer, all vectors are normalized to be in [0,1]. The second highlighter, as we call it fine highlighter, are composed of only a single fully-connected layer for simplicity. These two highlighters are designed to be serial to fit the serial deep-wise cross correlation and to make the features processed in a coarse-to-fine way.

As talked before, the class of the support images are the same as the objects in the query image. Specifically, the embeddings of the support images contain abundant information and then are produced to be factors which are used to do deep-wise cross correlation with the general features extracted from the query image. The deep-wise cross correlation is an operation which calculates the similarity between two units. Therefore, the operation is able to own the function that highlight the region of the object. Finally, through this operation, the general turns into object-specific.

### 3.2 Deep-wise Cross Correlation

After getting the general features from the feature extractor and the feature-exciting factors from the feature attention highlight module, a deep-wise cross correlation is implemented to combine them, like a connect unit. In short, these two feature maps with the same number of channels do the correlation operation channel by channel. Specifically, in Figure 3, for every channel of the general features, a convolution procedure whose convolutional kernel is the corresponding channel of the feature-exciting factors is applied on it.

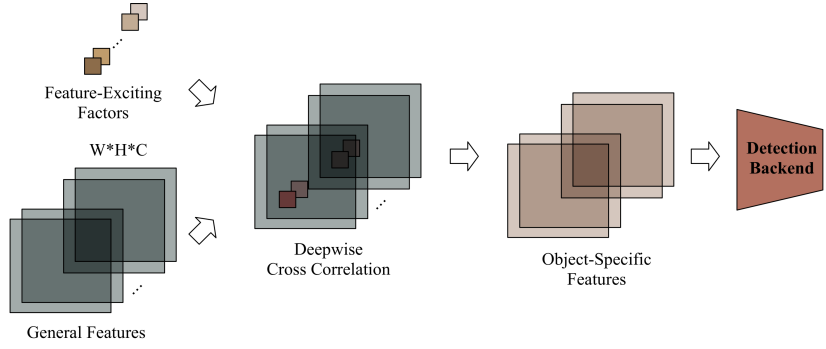

Figure 3. Deep-wise cross correlation.

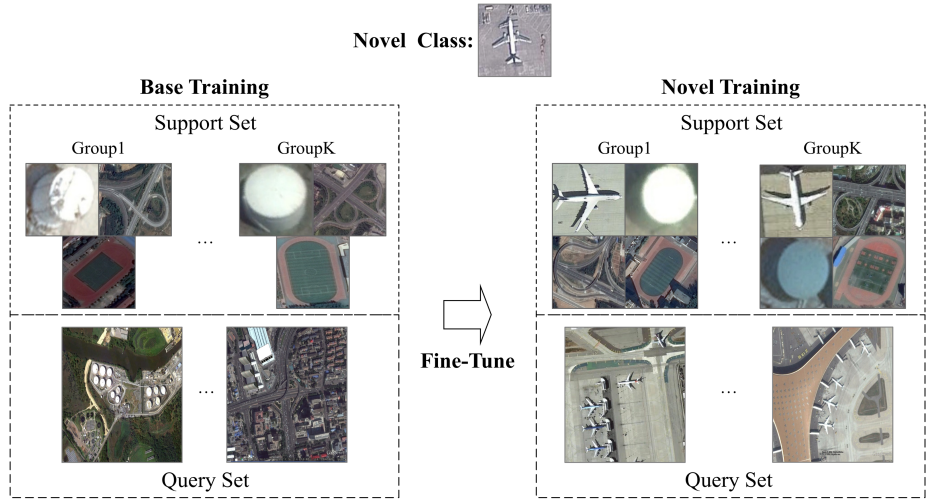

Figure 4. Our two-phase training scheme.

### 3.3 Training scheme

We adapt a two-phase training scheme in our work like transfer learning task to ensure that our model can obtain a generalized performance from few-shot settings. As shown in Figure 4, our training phase is divided into two parts: base training and novel fine-tuning.

In base training phase, only base classes with sufficient labeled data are available. As talked in introduction, the images in query set are directly from our dataset. The ones in support set are cropped from query set according to the annotation. In the second phase, the novel fine-tuning phase, the model is trained on not only base but also novel classes. The training data is the same like previous phase except that there are images from novel classes. Because there are only k annotations for each novel class, we still choose k annotations for each base class to get balance between them. Moreover, the procedure of the second phase is the same as the first one.

## 4. EXPERIMENT

### 4.1 Experiment Data Set

We evaluated the performance of our proposed networks on the dataset RSOD[18], which is a 4-class geospatial dataset for object detection. The 4 classes in RSOD are aircraft, oiltank, overpass and playground. Among these 4 classes, 1 class is randomly selected to become the novel class, while the rest 3 classes are selected to be base ones. Moreover, we evaluate the performance with 4 different base/novel splits for objectivity. In first phase, as for each base class, we randomly select 60% of the images to train our model then the rest ones for testing to verify the performance of our proposed approach. In novel fine-tuning, we only choose quite a small set of training images to make sure that there are only k annotated bounding boxes in each class, where k equals 1, 2, 3, 5 and 10.

### 4.2 Experiment Setup

Our model is compared with three baselines. The first baseline is to train the faster R-CNN with images from the base and novel classes together. Particularly, the number of annotations of each class fits the few-shot condition. In other words, there are only k annotations for not only novel classes but also base classes, where k equals 1, 2, 3, 5 and 10 here. We define this baseline as FRCN-few. The second baseline is the same like the first one except that there are abundant annotations in base classes and we define it as FRCN-joint. The rest one is trained with two training phases like ours. In detail, train the original faster R-CNN model the same as ours with only base classes in base training phase. Then in novel fine-tuning phase, the model is fine-tuned with base as well as novel classes. We define it as FRCN-ft. With the help of these three baselines, we can understand the important role that the feature attention highlight module plays in and in which condition our method can work better.

### 4.3 Performance Analysis

Table 1. Experimental results of few-shot object detection (AP of novel object). Specifically, "split 1" strands for taking aircraft as novel class and "split 2, 3, 4" stands for oiltank, overpass, playground as novel class respectively.

| Method/Shot | Split 1 | | | | | Split 2 | | | | | Split 3 | | | | | Split 4 | | | | |
|---|---|---|---|---|---|---|---|---|---|---|---|---|---|---|---|---|---|---|---|---|
| | 1 | 2 | 3 | 5 | 10 | 1 | 2 | 3 | 5 | 10 | 1 | 2 | 3 | 5 | 10 | 1 | 2 | 3 | 5 | 10 |
| FRCN-few | 0.05 | 0.23 | 1.30 | 3.03 | 9.09 | 0 | 0 | 0.61 | 1.44 | 46.96 | 1.65 | 4.66 | 6.19 | 11.18 | 20.66 | 12.16 | 37.29 | 49.61 | 50.59 | 66.02 |
| FRCN-joint | 0.19 | 9.09 | 9.09 | 9.91 | 17.04 | 9.09 | 9.39 | 11.23 | 17.86 | 59.84 | 1.32 | 3.79 | 10.98 | 25.52 | 32.88 | 9.35 | 19.15 | 37.96 | 31.73 | 76.12 |
| FRCN-ft | 2.27 | 6.26 | 10.76 | 16.05 | 31.34 | 13.09 | 14.66 | 48.04 | 57.94 | **72.54** | 3.99 | 8.74 | 16.78 | 31.65 | 51.84 | 28.84 | 34.95 | 52.18 | 75.99 | **90.58** |
| Ours | 9.1 | 10.6 | 15.0 | 20.2 | 43.5 | 16.54 | 34.8 | 51.6 | 60.92 | 71.27 | 9.09 | 11.36 | 21.01 | 41.4 | 59.63 | 33.54 | 50.5 | 61.6 | 77.1 | 88.98 |

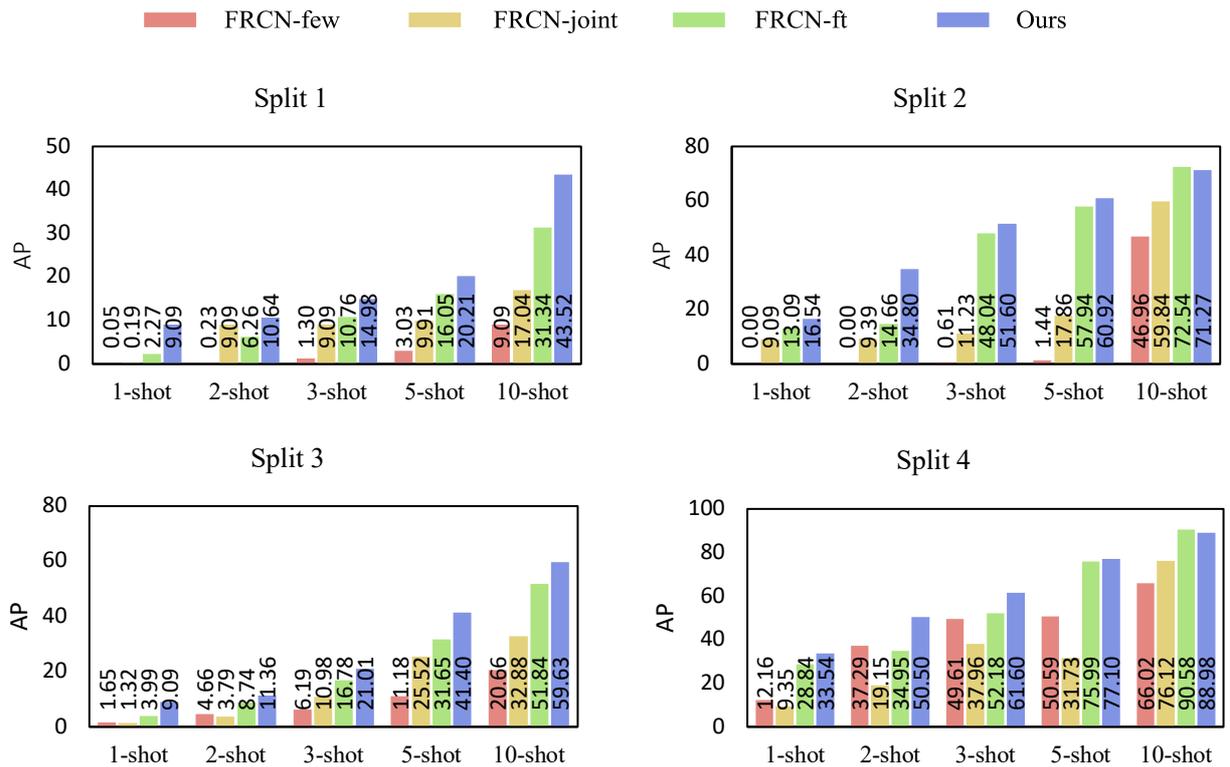

Figure 5. AP of novel object with histogram.

The results of our experiment on novel classes are in Table 1. First, in most conditions, our method outperforms the baselines which indicates the effectiveness and stability of our model. Moreover, the results of experiments on different splits of dataset show the generalized performance of our model when meeting different few-shot cases. Second, when the shots of novel classes increase, the AP of the detection improves. Furthermore, the performance of baseline 3 is better than the one of baseline 2 and the performance of baseline 2 is better than what is of baseline 1 in most cases. All these indicate that the information about a novel class implemented to detection the object of respective class comes from two parts: one is from the samples of the class itself and the other one is from the samples of other classes. When the information comes from other classes, the way to utilize it better is to fine-tune rather than training together. In short, the more useful information we extract, the better performance of detection of novel classes will show. As a consequence, the results demonstrate that our proposed method is able to obtain information from other classes in a more effective way when taking fine-tuning method.

On the other hand, as shown in Figure 5, when we focus on different objects, the difficulty of detecting them of different classes is different. For instance, the number of 10 or 10-shot of annotations of aircraft is quite not enough to detect it. However, as for oiltank or playground in the same experimental condition, it is much more enough. As a consequence, when we do few-shot learning research, the number of the object that we define as few may about to be different with the changes of the object we study.

## 5. CONCLUSION

In this work, we proposed a few-shot object detection model which is able to detect novel object with only a few annotated data in remote sensing images. A feature attention highlight module is proposed with a transfer training scheme that is adapted in the training procedure. With the help of the training scheme, the model is able to leverage knowledge from base classes and quickly adapt to novel classes. Experimental results based on the RSOD dataset verify that the effectiveness of our model when meeting few-shot detection cases.